\documentclass[11pt]{article}

\usepackage[final]{acl}

\usepackage{times}
\usepackage{latexsym}

\usepackage[T1]{fontenc}

\usepackage[utf8]{inputenc}

\usepackage{microtype}

\usepackage{inconsolata}

\usepackage{algorithm}
\usepackage{algpseudocode}   
\usepackage{graphicx}
\usepackage{enumitem}
\usepackage{amsmath}
\usepackage{booktabs}

\DeclareSymbolFont{txextras}{U}{txsyc}{m}{n}
\SetSymbolFont{txextras}{bold}{U}{txsyc}{bx}{n}
\DeclareMathSymbol{\txdiamond}{\mathord}{txextras}{113}
\DeclareMathSymbol{\txheart}{\mathord}{txextras}{114}

\usepackage[table]{xcolor}   

\definecolor{seeblue}{RGB}{222,235,247}  
\definecolor{basegray}{RGB}{242,242,242}

\usepackage{listings}
\usepackage[most]{tcolorbox}

\definecolor{promptbg}{RGB}{246,248,250}

\lstdefinestyle{prompt}{
    basicstyle=\ttfamily\scriptsize,
    backgroundcolor=\color{promptbg},
    breaklines=true,
    breakatwhitespace=true,
    frame=single,
    framesep=4pt,
    rulecolor=\color{gray!40},
    columns=fullflexible,
    keepspaces=true,
    xleftmargin=2pt,
    xrightmargin=2pt,
    aboveskip=4pt,
    belowskip=4pt
}

\newtcolorbox{baseresp}{
    colback=basegray,
    colframe=gray!55,
    fonttitle=\bfseries\small,
    title=Base response,
    boxrule=0.4pt,
    left=4pt,
    right=4pt,
    top=2pt,
    bottom=2pt,
    arc=2pt,
    fontupper=\small
}

\newtcolorbox{seeresp}{
    colback=seeblue,
    colframe=blue!30!gray,
    fonttitle=\bfseries\small,
    title=SEE response,
    boxrule=0.4pt,
    left=4pt,
    right=4pt,
    top=2pt,
    bottom=2pt,
    arc=2pt,
    fontupper=\small
}

\newtcolorbox{userbox}{
    colback=white,
    colframe=black!55,
    fonttitle=\bfseries\small,
    title=User request,
    boxrule=0.4pt,
    left=4pt,
    right=4pt,
    top=2pt,
    bottom=2pt,
    arc=2pt,
    fontupper=\small
}

\title{
Self-Evaluation Is Already There:
Eliciting Latent Judge Calibration in Base LLMs with Minimal Data
}

\author{
 \textbf{XiuYu Zhang}$^{\txheart}$\thanks{Equal contribution.},
 \textbf{Yi Shan}$^{\txdiamond}$\footnotemark[1],
 \textbf{Junfeng Fang}$^{\txheart}$\thanks{Corresponding author.},
 \textbf{Zhenkai Liang}$^{\txheart}$
\\
\\
 $^{\txheart}$National University of Singapore,
 $^{\txdiamond}$Beijing University of Technology
\\
 \small{
   xiuyu@comp.nus.edu.sg,
   yi.shan.bjut@gmail.com,
   fangjf@nus.edu.sg,
   liangzk@comp.nus.edu.sg
 }
}

\begin{document}

\maketitle

\begin{abstract}
Large language models are increasingly evaluated by other models, raising a natural question: can a model predict how a judge will score its own output? We find that the ability is largely present before any targeted training: prompted few-shot, a base model already predicts an external judge's multi-attribute quality scores on open-ended responses well above chance across three benchmarks. We introduce Self-Evaluation Elicitation (SEE), a method that surfaces this latent ability through a short cycle comprising a calibration-coupled reinforcement learning phase that improves the answer and predicts the judge, followed by a masked distillation phase that sharpens the prediction while leaving the answer untouched. From 160 unique examples, roughly $31\times$ fewer than a reinforcement learning baseline, SEE improves held-out calibration across three benchmarks while preserving answer quality. The elicited self-evaluation is sharply localized within the model's own token distribution and stable across judges it was never trained against, indicating a transferable notion of quality rather than a single judge's preference. These results reframe judge-aligned self-evaluation as a problem of elicitation rather than acquisition.
\end{abstract}

\section{Introduction}

Large language models (LLMs) are now routinely evaluated by other LLMs, and an LLM judge that rates qualities such as helpfulness and correctness has become a standard substitute for human annotation, both for benchmarking and as the reward signal in post-training~\citep{ouyang2022instructgpt, bai2022constitutional, zheng2023judging, lee2024rlaif}.
A natural question follows from this trend: if a judge will score a model's output, can the model anticipate that score for its own output?
%
%
The ability would be useful, since a model that predicts how it will be judged can rerank its own samples, defer when it expects a low score, or escalate a difficult prompt to a stronger model, none of which requires querying the judge at inference time.

An LLM that could do this reliably would be valuable, and recent work has indeed trained models to predict a reward signal in a relatively narrow setting.
These methods operate on verifiable tasks such as mathematics and reasoning, where the target is a scalar measure of correctness relative to a known answer, and the predicted score serves primarily to improve the final output~\citep{damani2025rlcr, fei2025pcl, yang2025laser}.
It remains unclear (1) whether a model can predict an external judge's multi-attribute quality scores in the open-ended setting, where no verifiable answer exists, and (2) how much of this ability a base model already possesses before any targeted training.

We begin our exploration from the second question, and its answer reshapes the first.
We discover that the base model already approximates the judge to a substantial degree. 
Prompted few-shot in our scoring format, Qwen3-4B-Base~\citep{yang2025qwen3technicalreport} predicts an external judge's five attributes~\citep{wang2024helpsteer} with a nonlinear calibration of $0.50\sim 0.70$ across three benchmarks, well above random guessing, despite never having been trained to do so specifically.
The predictions are noisy and often overconfident, but they track the judge far more closely than chance would, suggesting that the representation needed for self-evaluation is developed during pretraining and only needs to be surfaced.
This interpretation is consistent with a broader body of evidence.
Base models already carry a usable signal about whether their own answers are roughly correct~\citep{kadavath2022know}.
Small data alignment and reasoning work indicates that post-training largely surfaces capabilities already present in pre-training rather than adding new ones~\citep {zhou2023lima,muennighoff2025s1,ye2025limo}, and reinforcement learning (RL) has been shown in several settings to elicit behavior that the base model can already produce rather than create it~\citep{yue2026does,shao2025spurious, zhang2026alphaalign}.

If the ability is already present, surfacing it should require little data, and we accordingly replace the usual large training run with a short cycle of two alternating phases.
First, a brief RL phase improves the answer in response to the judge's reward. 
Second, a subsequent distillation phase takes the rollouts collected during the first RL phase and trains on the judge's actual scores, with the loss restricted to the self-evaluation tokens, leaving the answer itself unchanged. 
The second phase thus amounts to on-policy distillation of the judge into the self-evaluation channel~\citep{agarwal2024gkd}.
Repeating this cycle a small number of times consumes roughly 160 unique samples in total and noticeably surfaces the self-evaluation ability.

RL optimizes the whole response, thereby improving the answer and the self-evaluation at once, while the masked distillation phase corrects only the self-evaluation, re-anchoring its scores to the judge on the distribution the model now produces. 
Because that correction is confined to the
self-evaluation tokens, calibration is sharpened without disturbing the answer. 
After 15 cycles, the self-prediction error decreases by $0.25\sim 0.66$ mean absolute error (MAE) across all benchmarks we evaluate, calibration improves correspondingly, and answer quality remains unchanged or slightly improves.
This result outperforms pure RL with two full epochs of substantially more training data. 
The improvement appears where the latent-ability account predicts it should, i.e., in the self-evaluation tokens we train, and not in the answer distribution we leave untouched.

Our contributions can be summarized as follows:
\begin{enumerate}[nosep,leftmargin=*]
    \item \textbf{Reframing self-evaluation as a problem of elicitation rather than acquisition.}
    We show that base LLMs already approximate a multi-attribute judge's scores, and we measure how much of this ability is present before training, thereby reframing the task of self-evaluation.
    \item \textbf{Light-weight cyclic RL-and-SFT procedure to surface latent ability.} 
    We introduce Self-Evaluation Elicitation (SEE), a method that alternates reinforcement learning with a masked distillation phase and extracts ability from 160 unique training examples, roughly $31\times$ fewer than the reinforcement learning baseline, while improving held-out calibration across three benchmarks and preserving answer quality.
    \item \textbf{Robust elicited self-evaluation.} The judge's score falls within the model's top-5 predicted tokens at high rates, and the quality and calibration gains persist when responses are scored by held-out judges rather than the training judge. A \texttt{Llama-3-8B-Base} replication extends the evidence across model families, and component ablations isolate the roles of RL, masking, and stratified sampling.
\end{enumerate}

\section{Related Work}

\paragraph{Self-evaluation as an RL signal.}

Recent works augment reinforcement learning (RL) so that the trained model emits a score predicting the reward it will receive alongside its response to the given prompt. 
RLCR adds a Brier-score confidence term to the reward, training the model to output a calibrated probability estimate of the correctness of its answer, and shows the construction holds for any bounded proper scoring rule~\citep{damani2025rlcr}.
PCL has the model reproduce the rule-based reward it was optimized against, then discard this self-assessment at inference to keep generation unchanged~\citep{fei2025pcl}.
LaSeR signs a last-token self-reward with a verifier signal, allowing the model to score its own reasoning~\citep{yang2025laser}.
Across these methods, the prediction target is verifiable correctness, a scalar defined against a known answer, and the studies are conducted on mathematics, reasoning, or coding, where such an answer exists~\citep{bian2026expthink,he2026order}.
Of the three, RLCR is closest to our reward design: its proper-scoring-rule penalty on the gap between predicted and true scores is nonlinear, as ours is, whereas PCL's consistency reward is linear, and LaSeR's alignment is an auxiliary loss rather than a reward term.
We adopt RLCR as our principal baseline for our setting.
With our Self-Evaluation Elicitation (SEE), we predict an external judge's score along several quality attributes~\citep{wang2024helpsteer, tan2026papo} on open-ended prompts where no verifiable answer is available.

\paragraph{Eliciting latent capability.}

A growing number of works argue that post-training surfaces abilities a base model already holds rather than installing new ones.
Base models can already estimate whether their own answers are correct to a certain extent~\citep{kadavath2022know}.
Alignment on a thousand or so examples recovers most of the quality of large-scale instruction tuning~\citep{zhou2023lima}, and completion-level reasoning can be elicited from comparably small sets, which their authors frame explicitly as knowledge elicitation rather than acquisition~\citep{muennighoff2025s1, ye2025limo}.
RL, too, has been shown to increase the probability of behavior the base model can already produce rather than to extend its reach beyond it~\citep{yue2026does, shao2025spurious}, and a few hundred RL steps can surface latent safety behavior that pretraining had already installed~\citep{zhang2026alphaalign}.
This evidence concerns task accuracy, self-knowledge about correctness, and safety. 
Whether the same pattern holds for judge-aligned, multi-attribute self-evaluation, predicting how an external judge will rate an open-ended response has not been examined.
We provide that measurement and build a method on top of it. 

\begin{figure*}[t]
  \centering
  \includegraphics[width=\textwidth]{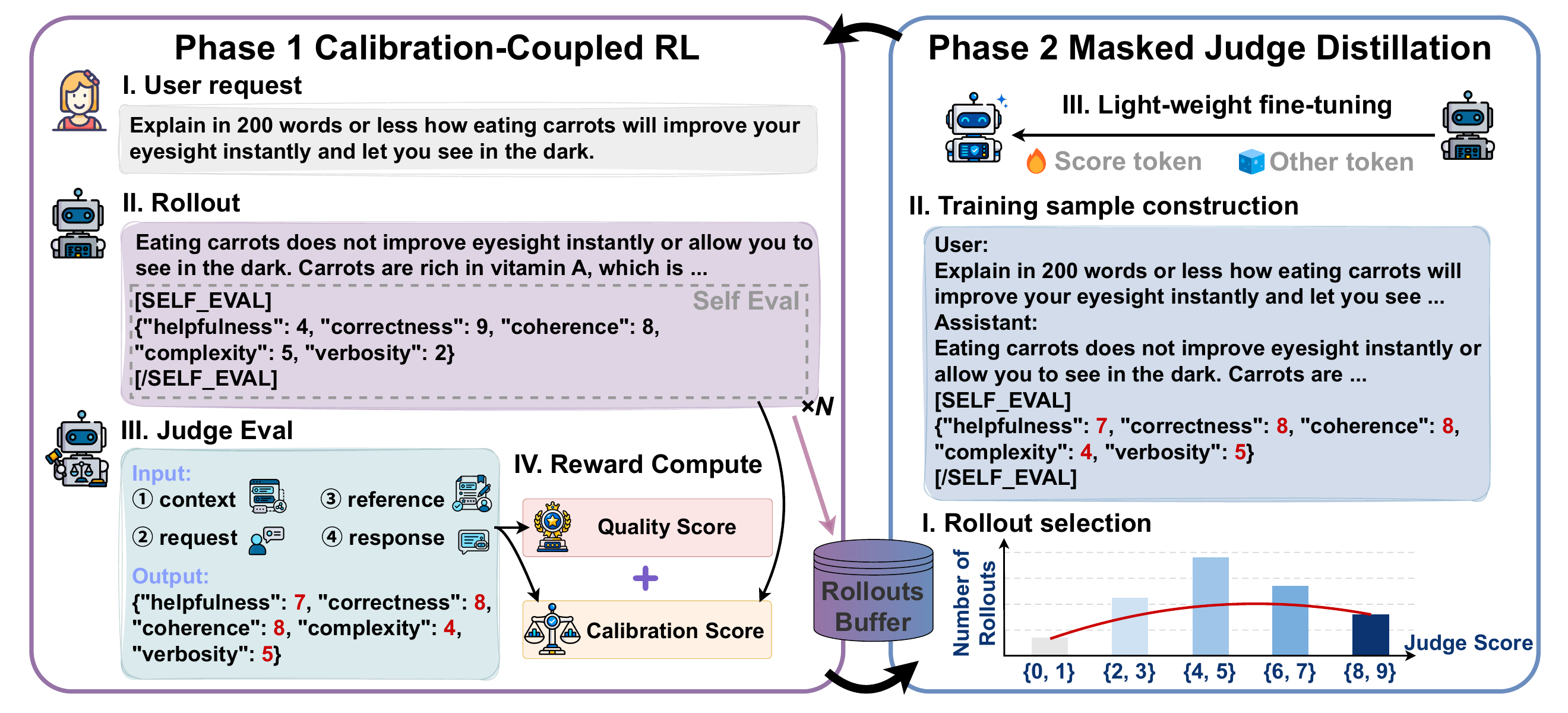}
  \caption{Overview of the SEE cycle. \textbf{Phase 1, Calibration-Coupled RL:} the policy answers a prompt and appends an inline \texttt{[SELF\_EVAL]} block of five attribute scores (\emph{Self Eval}); an external judge scores the same response on the same attributes; the reward combines a quality term over the three evaluative attributes with a calibration term over all five (Eq.~\ref{eq:reward}). Each rollout, with its self-scores and judge scores, is written to a buffer ($\times N$). \textbf{Phase 2, Masked Judge Distillation:} rollouts are selected from the buffer by stratified round-robin over attribute-score bins, training samples are built by filling the \texttt{[SELF\_EVAL]} block with the judge's scores, and the model is fine-tuned with the loss applied only to the score tokens and not to the rest of the response. The two phases alternate.}
  \label{fig:method}
\end{figure*}

\paragraph{Alternating supervised and reinforcement learning.}

Several methods improve a model by alternating RL with supervised fine-tuning (SFT) on the model's own rollouts. 
ReST, ReST\textsuperscript{EM}, RAFT, and STaR generate samples from the current policy, keep those that score well under a reward, and fine-tune the model on the survivors~\citep{gulcehre2023rest, singh2024restem, dong2023raft, zelikman2022star}.
On-policy distillation follows a similar recipe, training a student on its own generations from a teacher to close the gap between the training and inference distributions~\citep{agarwal2024gkd}.
In both cases, the supervised signal falls on the answer, and the rollouts are filtered to those that the reward already favors, so fine-tuning refines the model's better answers.
Our supervised phase, i.e., the distillation phase of SEE, does neither. 
It leaves the answer tokens untouched, applies its loss only to the self-evaluation block, and retains rollouts across the full range of scores, so that the model learns to predict both low and high judge scores.
The phase is therefore better understood as on-policy distillation of the judge into a self-evaluation channel, run alongside, but kept separate from, the RL that improves the answer.

\paragraph{LLM judges and multi-attribute reward.}
Using one language model to score another's output is now standard practice: strong judges agree with human preference about as often as humans agree with each other~\citep{zheng2023judging}, and datasets such as HelpSteer2 rate responses along separate attributes of helpfulness, correctness, coherence, complexity~\citep{wang2024helpsteer}.
We adopt this multi-attribute view but turn it inward. Rather than training an external reward model to generate scores~\citep{fu2025learning}, we ask the policy to predict the judge's scores for its own outputs and study how well it does.
\section{Method}
\label{sec:method}


%
We introduce \textbf{Self-Evaluation Elicitation (SEE)}, a method that surfaces a model's latent ability to predict that judge's scores for its own outputs, while also improving the outputs themselves.
SEE applies to the open-ended setting, where a model answers prompts that have no verifiable ground-truth answer and a multi-attribute judge scores its responses.
The method alternates two phases over a single model.
The first phase, \emph{Calibration-Coupled RL}, jointly optimizes response quality and calibration.
During each rollout, the model generates an answer and predicts the scores the judge will assign; the reward reflects both the answer quality and the accuracy of these predictions.
The second phase, \emph{Masked Judge Distillation}, reuses these rollouts to train the model on the scores the judge actually assigns.
Because the loss is applied only to the self-evaluation tokens, this phase does not directly optimize the generated answer.
Repeating the two phases, the \emph{SEE cycle}, re-grounds the prediction to the judge as the answer distribution moves. 
SEE adds no separate reward model and updates only the self-evaluation tokens during distillation, and we show that this suffices to elicit the capability from little data.

\subsection{Calibration-Coupled RL}
\label{sec:method:rl}

Given a prompt, the model produces a response followed by a single inline self-evaluation block, delimited by \texttt{[SELF\_EVAL]} and
\texttt{[/SELF\_EVAL]}, containing integer scores on a $0$--$9$ scale for the five HelpSteer2 attributes of helpfulness, correctness, coherence, complexity, and verbosity~\citep{wang2024helpsteer}.
An external judge scores the same response on the same five attributes. 
Let $s$ be the model's self-scores and $j$ be the judge's scores; the reward is a quality term and a calibration term when the self-evaluation block is well-formed, and a fixed penalty otherwise,
\begin{equation}
\begin{aligned}
r &=
\begin{cases}
-1, & \text{if malformed}, \\
w_q q(j) + w_c c(s,j), & \text{otherwise},
\end{cases} \\
q(j) &= \tfrac{1}{3}\sum_{a \in \mathcal{E}} \tfrac{j_a}{9}, \\
c(s,j) &= \Big(1 - \tfrac{1}{9}\,\mathrm{MAE}(s,j)\Big)^{\gamma}.
\end{aligned}
\label{eq:reward}
\end{equation}
where $\mathcal{E}=\{\mathrm{hlp,cor,coh}\}$, a block is well-formed only if it parses to integer scores in $[0,9]$ for all five attributes, $\mathrm{MAE}(s,j)$ is the mean absolute error over those five attributes, $w_q$ and $w_c$ weight the two terms, and $\gamma$ controls how sharply large disagreements are penalized.
The penalty makes the reward fail closed: a response whose self-evaluation cannot be parsed receives the minimum reward regardless of how good the answer is, which pressures the model to keep the block well-formed.
We optimize this reward with GRPO~\citep{shao2024GRPO} over the full response, with no special handling of the self-evaluation tokens, so the calibration term is what shapes those tokens toward the judge while the quality term shapes the answer.
We write each well-formed rollout, the response together with its self-scores and the judge's scores, to a buffer for use in the second phase; rollouts that received the penalty are discarded.

The quality term and the calibration term read different attributes by design. 
Quality averages only the three evaluative attributes, the ones a good answer should maximize, whereas calibration is measured over all five, including the descriptive attributes of complexity and verbosity that the model should predict accurately.
RL on quality alone would improve the answer but leave the self-evaluation unconstrained, free to drift toward an uninformative, overconfident constant.

We use a nonlinear exponent $\gamma > 1$ to place more pressure on large disagreements.
A linear calibration reward treats a one-point error and a four-point error as differing only in degree, so a model can collect most of the reward by predicting near the judge's mean.
Raising the agreement to a power amplifies large gaps, pushing self-scores towards the judge's values across the score range rather than towards its center.

\subsection{Masked Judge Distillation}
\label{sec:method:sft}

The second phase turns the buffered rollouts into supervised targets.
Because malformed rollouts were discarded during the first phase, every rollout in the buffer already carries a parseable self-evaluation block, so the format constraint is enforced once rather than separately in each phase.
We select rollouts by a stratified round-robin over a grid of $5 \times 5 = 25$ cells, one axis the five attributes and the other five score bins ($\{0,1\}, \{2,3\}, \dots, \{8,9\}$): on each pass we shuffle the cell order and draw one not-yet-selected sample from each non-empty cell, repeating until we reach \texttt{SFT\_MAX\_SAMPLES} and topping up with a random draw from the remainder if the grid empties first. 
For each selected rollout, we build a training sample from the prompt and the model's own response, fill the self-evaluation block with the \emph{judge's} scores rather than the model's, and fine-tune with the loss applied only to the tokens inside the self-evaluation block.

We include this phase to target two problems. 
The calibration reward in Equation~\ref{eq:reward} is a single scalar per rollout and a slow teacher. Supervising directly on the judge's five scores is a denser, faster signal that sharpens self-evaluation on far less data. 
At the same time, fine-tuning the whole response on judge-scored rollouts would distort the answer distribution towards whatever the judge happened to favor, undoing the work of the first phase.
Restricting the loss to the self-evaluation tokens removes those risks, i.e., the update reaches the prediction and nothing else.

Since the responses are the model's own current outputs, the distillation grounds the self-evaluation to the judge on exactly the distribution the model produces, not on a fixed external corpus~\citep{agarwal2024gkd}.
The balanced selection matters for the same reason the non-linear reward does.
A buffer of open-ended responses is dominated by mid-range judge scores, so sampling it directly would teach the model to predict the middle well and the extremes poorly. Resampling towards even coverage of the $25$ cells gives the rare low and high scores enough weight to be learned. 

\subsection{The SEE Cycle}
\label{sec:method:cycle}

Calibration-Coupled RL continuously shifts the answer distribution, so a self-evaluation distilled only before RL would describe responses the model no longer produces.
Running the two phases in alternation re-grounds the prediction to the judge after each round of answer improvement.
We therefore interleave them, alternating a phase of RL with a single distillation pass and repeating for several cycles.
RL improves the answer and the self-evaluation together, since it optimizes the full response under a reward that rewards both; the distillation pass then corrects only the self-evaluation, because its loss is confined to those tokens, sharpening the prediction against the judge's exact scores without disturbing the answer RL produced. 
Thus, the two updates reinforce rather than compete, and answer quality and self-evaluation accuracy improve across cycles; Section~\ref{sec:exp:components} isolates the contribution of each component.

\section{Experiments}
\label{sec:experiments}

\begin{table*}[t]
\centering
\small
\setlength{\tabcolsep}{5pt}
\begin{tabular}{l cc>{\columncolor{seeblue}}c cc>{\columncolor{seeblue}}c cc>{\columncolor{seeblue}}c}
\toprule
& \multicolumn{3}{c}{Response Win-rate} & \multicolumn{3}{c}{Quality} & \multicolumn{3}{c}{Calibration} \\
\cmidrule(lr){2-4}\cmidrule(lr){5-7}\cmidrule(lr){8-10}
Benchmark & Qwen3-4B & \shortstack{Adapted\\RLCR} & SEE & Qwen3-4B & \shortstack{Adapted\\RLCR} & SEE & Qwen3-4B & \shortstack{Adapted\\RLCR} & SEE \\
\midrule
LC AlpacaEval 2.0    & 0.500 & 0.534 & \textbf{0.592} & 0.788 & 0.789 & \textbf{0.792} & 0.702 & 0.716 & \textbf{0.746} \\
Arena-Hard-Auto v2.0 & 0.500 & 0.502 & \textbf{0.518} & 0.568 & 0.569 & \textbf{0.575} & 0.517 & 0.547 & \textbf{0.609} \\
WildBench v2         & 0.500 & 0.537 & \textbf{0.581} & 0.560 & 0.561 & \textbf{0.567} & 0.504 & 0.541 & \textbf{0.609} \\
\bottomrule
\end{tabular}
\caption{Open-ended benchmarks. Response win-rate is a pairwise GPT-5.4
preference against the base model's answers; quality and calibration are the
judge's five-attribute scores as defined in Section~\ref{sec:exp:setup}. SEE is
best on every benchmark and metric.}
\label{tab:main}
\end{table*}

\begin{figure*}[t]
\centering
\includegraphics[width=\textwidth]{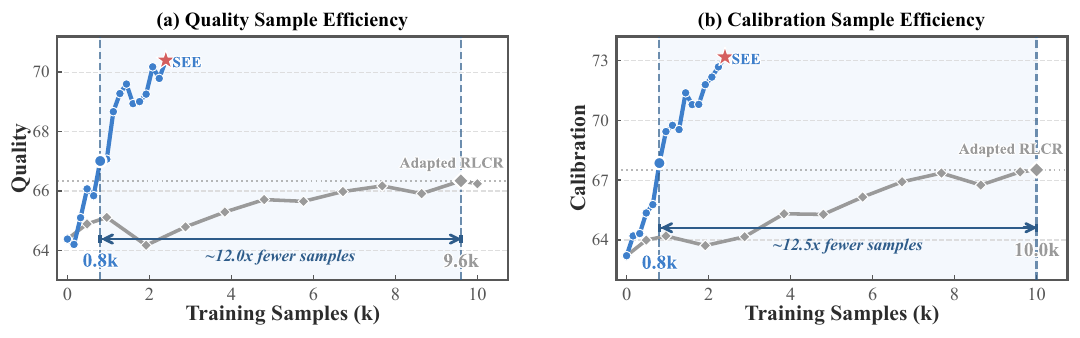}
\caption{Sample efficiency. Quality (left) and calibration (right) against sample-passes for SEE and Adapted RLCR. SEE reaches the baseline's final scores after $\sim$0.8k sample-passes ($\sim$$12\times$ fewer) and keeps improving. The $x$-axis counts sample-passes, not unique examples; on unique examples the gap is $\sim$$31\times$.}
\label{fig:efficiency}
\end{figure*}
 
\subsection{Experimental setup}
\label{sec:exp:setup}
We train \texttt{Qwen3-4B-Base} with SEE and replicate the full comparison on \texttt{Llama-3-8B-Base}. The evaluation follows an evidence chain: we first test whether an untrained Qwen already predicts the judge (Section~\ref{sec:exp:latent}); then whether SEE improves quality and calibration efficiently (Section~\ref{sec:exp:main}); whether the result transfers across base models and judges (Section~\ref{sec:exp:robust}); and finally which components produce the gains and what the resulting calibration looks like (Sections~\ref{sec:exp:components} and~\ref{sec:exp:attr}).

\paragraph{Judge.}
A single training judge, GPT-5.4, supplies all supervision and the primary evaluation.
During training, it scores each response on the five HelpSteer2 attributes~\citep{wang2024helpsteer}, and these scores drive the reward in Equation~\ref{eq:reward}.
At evaluation, it produces the same five-attribute scores, from which we compute quality and calibration, and on the open-ended benchmarks it acts as a pairwise referee between two models' responses. We separately re-score outputs with two held-out judges in Section~\ref{sec:exp:robust}.

\paragraph{Benchmarks.}
We report on HelpSteer2 validation (unique prompt subset) and on three open-ended instruction-following benchmarks: LC AlpacaEval~2.0~\citep{dubois2025lengthcontrol}, Arena-Hard-Auto~v2.0~\citep{li2025from}, and WildBench~v2~\citep{lin2024wildbenchbenchmarkingllmschallenging}.
HelpSteer2 validation carries the judge's recorded five-attribute scores, so we use it to measure score-level agreement; the other three tests transfer to standard open-ended evaluation.

\paragraph{Metrics.}
Only responses with the correct \texttt{[SELF\_EVAL]} format are used for the calculation of the following metrics.
\emph{Quality} is the judge's mean over the three evaluative attributes (helpfulness, correctness, coherence), normalized to $[0,1]$.
\emph{Calibration} measures agreement between the model's self-scores and the judge's scores using the same nonlinear form as the training reward, $\big(1-\tfrac{1}{9}\mathrm{MAE}(s,j)\big)^{\gamma}$ over all five attributes.
The reported metric matches the calibration term optimized by both SEE and Adapted RLCR.
We report two kinds of win-rate, both against the base model and both as $(\text{wins}+0.5\cdot\text{ties})/n$.
A \emph{score win-rate} compares the policy's and the base model's recorded quality (or calibration) scores sample by sample, with no additional judge call.
A \emph{response win-rate} instead asks the judge for a single pairwise preference between the two models' answers without using the five-attribute scoring.
For self-evaluation localization, we report \emph{top-5 accuracy}: at each score position, we rank the ten score-digit tokens by the model's logits and count a hit when the judge's score token lies in the top five, averaged over all attributes and examples. Choosing five of ten tokens uniformly gives a 50\% random floor.

\paragraph{Baseline.}
RLCR augments an RL correctness reward with a Brier-score calibration term and shows the construction holds for any bounded proper scoring rule~\citep{damani2025rlcr}; it targets verifiable tasks, where correctness is checked against ground truth, which does not hold in open-ended judging.
We therefore compare against \emph{Adapted RLCR}: RLCR's calibration-coupled RL moved to our setting, with its scalar binary-correctness target replaced by our multi-attribute judge-score calibration term.
Adapted RLCR is a single-phase run: it uses the same reward, prompts, judge, and number of rollouts as SEE's Calibration-Coupled RL phase but omits the Masked Judge Distillation phase, making it the closest single-phase analog to SEE.
The reported runs differ in batch size (48 for Adapted RLCR, 16 for SEE), so we additionally train SEE at batch size 48 as a matched control (Section~\ref{sec:exp:components}).

\begin{table}[t]
\centering
\small
\setlength{\tabcolsep}{5pt}
\begin{tabular}{lcccc}
\toprule
HelpSteer2 Val & Quality & Calib. & \shortstack{Win-rate\\Quality} & \shortstack{Win-rate\\Calib.} \\
\midrule
Qwen3-4B          & 0.644 & 0.632 & 0.500 & 0.500 \\
Adapted RLCR      & 0.662 & 0.675 & 0.570 & 0.617 \\
\rowcolor{seeblue}
SEE               & \textbf{0.704} & \textbf{0.731} & \textbf{0.671} & \textbf{0.700} \\
\bottomrule
\end{tabular}
\caption{HelpSteer2 validation. Quality and calibration are defined in Section~\ref{sec:exp:setup}; win-rates are score win-rates against the base model (no extra judge call). SEE uses 160 unique examples, Adapted RLCR
$\sim$5{,}000.}
\label{tab:helpsteer}
\end{table}

\paragraph{Training budget.}
SEE runs the cycle for 15 rounds across 160 unique examples, for 2{,}400 total sample-passes, for both Qwen and Llama.
Adapted RLCR trains for two epochs over roughly 5{,}000 unique examples, about 10{,}000 sample-passes.
We set $w_q=0.7$, $w_c=0.3$, and $\gamma=2$, and optimize with GRPO~\citep{shao2024GRPO}.

\begin{table*}[t]
\centering
\small
\setlength{\tabcolsep}{10pt}
\begin{tabular}{l cc cc>{\columncolor{seeblue}}c cc>{\columncolor{seeblue}}c}
\toprule
& \multicolumn{2}{c}{Win-rate} & \multicolumn{3}{c}{Quality} & \multicolumn{3}{c}{Calibration} \\
\cmidrule(lr){2-3}\cmidrule(lr){4-6}\cmidrule(lr){7-9}
Benchmark & \shortstack{Adapted\\RLCR} & SEE & Base & \shortstack{Adapted\\RLCR} & SEE & Base & \shortstack{Adapted\\RLCR} & SEE \\
\midrule
HelpSteer2 val       & 0.720 & \textbf{0.865} & 0.457 & 0.589 & \textbf{0.671} & 0.501 & 0.597 & \textbf{0.749} \\
LC AlpacaEval 2.0    & 0.578 & \textbf{0.630} & 0.581 & 0.661 & \textbf{0.669} & 0.580 & 0.629 & \textbf{0.682} \\
Arena-Hard-Auto v2.0 & 0.548 & \textbf{0.568} & 0.327 & 0.352 & \textbf{0.369} & 0.380 & 0.384 & \textbf{0.511} \\
WildBench v2         & 0.594 & \textbf{0.635} & 0.328 & 0.393 & \textbf{0.405} & 0.369 & 0.417 & \textbf{0.531} \\
\bottomrule
\end{tabular}
\caption{Replication on \texttt{Llama-3-8B-Base}. Base win-rate is 0.500 by construction and omitted. HelpSteer2 uses quality score win-rate; the open-ended benchmarks use pairwise response win-rate. SEE exceeds Adapted RLCR and the base model on every metric and benchmark.}
\label{tab:llama-main}
\end{table*}

\subsection{Evidence before training}
\label{sec:exp:latent}

Before asking whether SEE improves self-evaluation, we test whether the base model contains a usable signal to elicit. Prompted few-shot in the target format, \texttt{Qwen3-4B-Base} reaches 0.63 calibration on HelpSteer2 validation and 0.50--0.70 across the three open-ended benchmarks (Tables~\ref{tab:helpsteer} and~\ref{tab:main}).
It also places the judge's score among its five most probable score tokens 77.07\% of the time on HelpSteer2 validation, 27.07 points above the 50\% random floor (Table~\ref{tab:topk}).
Because these measurements precede targeted training, they establish the premise for the remaining experiments: Qwen already carries a judge-predictive signal, and SEE is tested as a way to sharpen it.

\subsection{SEE improves quality and calibration efficiently}
\label{sec:exp:main}
\label{sec:exp:efficiency}

Table~\ref{tab:helpsteer} first evaluates score-level agreement on HelpSteer2 validation. SEE improves both quality and calibration over the base model and Adapted RLCR, and wins the majority of per-sample comparisons against the base model on both scores. With 160 unique examples, SEE reaches 0.7312 calibration, versus 0.6752 for Adapted RLCR trained on roughly $31\times$ more unique examples.

The same ordering holds on the three open-ended benchmarks (Table~\ref{tab:main}). SEE is best on response win-rate, quality, and calibration throughout, with the clearest separation in calibration: on WildBench~v2, calibration rises from 0.5040 for the base model to 0.6088 for SEE, while Adapted RLCR reaches 0.5414. The SEE--Adapted RLCR quality gains are smaller, $+0.003$ to $+0.006$, while calibration and win-rate improve substantially.

\paragraph{Sample efficiency.}
The unique-example comparison understates the training-time gap because SEE revisits its small pool across cycles. Figure~\ref{fig:efficiency} therefore plots both methods against sample-passes. SEE reaches Adapted RLCR's final quality and calibration after about 0.8k sample-passes, roughly $12\times$ fewer than the baseline's $\sim$9.6--10k, and continues to improve. Thus, SEE is efficient both in unique data ($31\times$ fewer examples) and in the number of processed examples required to reach the baseline ($12\times$ fewer sample-passes).

\subsection{Generalization across models and judges}
\label{sec:exp:llama}
\label{sec:exp:robust}

We test generalization along two axes: the base model and the judge that supplies its reward.

\paragraph{Across base models.}
We apply the same 160-prompt, 15-cycle recipe to \texttt{Llama-3-8B-Base}~\citep{meta2024llama3}. Table~\ref{tab:llama-main} preserves SEE $>$ Adapted RLCR $>$ base for every metric and benchmark, extending the result to a second family and scale. Llama begins from a weaker signal, with base calibration of 0.501 on HelpSteer2 validation versus Qwen's 0.632, and still benefits consistently from SEE. Llama-3-8B also predates HelpSteer2, providing an additional test of transfer beyond Qwen.

\paragraph{Across judges.}
\begin{table*}[t]
\centering
\setlength{\tabcolsep}{4pt}
\resizebox{\textwidth}{!}{%
\begin{tabular}{l cc>{\columncolor{seeblue}}c cc>{\columncolor{seeblue}}c cc>{\columncolor{seeblue}}c cc>{\columncolor{seeblue}}c}
\toprule
& \multicolumn{6}{c}{Claude Sonnet 4.6} & \multicolumn{6}{c}{Gemini 3.1 Flash-Lite} \\
\cmidrule(lr){2-7}\cmidrule(lr){8-13}
& \multicolumn{3}{c}{Quality} & \multicolumn{3}{c}{Calibration}
& \multicolumn{3}{c}{Quality} & \multicolumn{3}{c}{Calibration} \\
\cmidrule(lr){2-4}\cmidrule(lr){5-7}\cmidrule(lr){8-10}\cmidrule(lr){11-13}
Benchmark
& \shortstack{Qwen\\3-4B} & \shortstack{Adapt.\\RLCR} & SEE
& \shortstack{Qwen\\3-4B} & \shortstack{Adapt.\\RLCR} & SEE
& \shortstack{Qwen\\3-4B} & \shortstack{Adapt.\\RLCR} & SEE
& \shortstack{Qwen\\3-4B} & \shortstack{Adapt.\\RLCR} & SEE \\
\midrule
HelpSteer2 val  & 0.575 & 0.581 & \textbf{0.599} & 0.585 & 0.621 & \textbf{0.670} & 0.682 & 0.691 & \textbf{0.718} & 0.651 & 0.670 & \textbf{0.709} \\
AlpacaEval 2.0  & 0.637 & 0.643 & \textbf{0.660} & 0.609 & 0.645 & \textbf{0.704} & 0.828 & 0.829 & \textbf{0.843} & 0.650 & 0.655 & \textbf{0.690} \\
Arena-Hard v2.0 & 0.464 & 0.470 & \textbf{0.483} & 0.443 & 0.497 & \textbf{0.551} & 0.631 & 0.638 & \textbf{0.658} & 0.522 & 0.556 & \textbf{0.612} \\
WildBench v2    & 0.465 & 0.471 & \textbf{0.488} & 0.430 & 0.500 & \textbf{0.564} & 0.633 & 0.637 & \textbf{0.649} & 0.524 & 0.560 & \textbf{0.620} \\
\bottomrule
\end{tabular}}
\caption{Cross-judge generalization. The Qwen models are trained against GPT-5.4 and re-scored by two held-out judges. SEE $>$ Adapted RLCR $>$ base for every judge, benchmark, and metric; absolute scores change with the judge, but the ordering persists.}
\label{tab:crossjudge}
\end{table*}

Because GPT-5.4 supplies the training signal, we re-score the Qwen models' outputs with Claude Sonnet 4.6 and Gemini 3.1 Flash-Lite. Table~\ref{tab:crossjudge} preserves SEE $>$ Adapted RLCR $>$ base in quality and calibration across every held-out judge and benchmark, despite shifts in absolute score. Llama shows the same ordering for quality and calibration in all eight held-out judge--benchmark conditions and for win-rate in seven of eight (Appendix~\ref{app:additional:crossjudge}); under Gemini on Arena-Hard, SEE trails Adapted RLCR by 0.005. Together, the Llama replication and held-out-judge evaluation extend the main ordering across model families and evaluators.

\subsection{Components and controls}
\label{sec:exp:components}

We isolate the contribution of each SEE component and match the RL batch size used by Adapted RLCR.

\paragraph{Component ablations.}
\begin{figure}[t]
\centering
\includegraphics[width=\columnwidth]{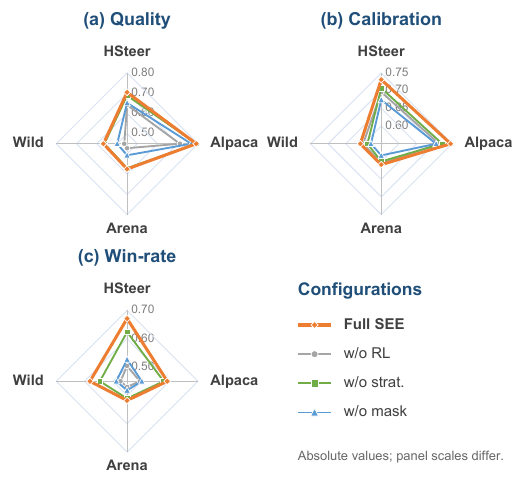}
\caption{Qwen3-4B component ablations across four benchmarks. Each radar reports absolute values; panel scales differ. Full numerical results are in Appendix~\ref{app:additional:ablations}.}
\label{fig:ablation-radar}
\end{figure}

Figure~\ref{fig:ablation-radar} removes each component in turn. Without RL weight updates, quality drops by 0.062--0.103 and win-rate approaches parity, showing that RL carries answer quality. Removing the score-token mask lowers quality on all four benchmarks and similarly pulls win-rate towards parity, showing that masking preserves the RL-improved answer. Replacing stratified selection with random buffer sampling leaves quality nearly unchanged but lowers calibration by 0.020--0.024 on three benchmarks and by 0.009 on Arena-Hard. Appendix~\ref{app:additional:ablations} reports all values.

\paragraph{Batch-matched control.}
We retrain SEE with RL batch size 48, matching Adapted RLCR. Matched-batch SEE retains calibration gains of 0.025--0.047 and win-rate gains of 0.020--0.103 over Adapted RLCR on every benchmark, while moving SEE from batch 16 to 48 changes any metric by at most 0.022 (Appendix~\ref{app:additional:batch}). The matched control preserves the reported separation.

\subsection{Calibration behavior}
\label{sec:exp:attr}

Aggregate calibration can conceal where the judge-aligned signal sits in the model's distribution and whether gains concentrate on particular scores or attributes. We examine all three views.

\paragraph{Token localization.}
\begin{table}[t!]
\centering
\setlength{\tabcolsep}{4pt}
\resizebox{\columnwidth}{!}{%
\begin{tabular}{lcccc}
\toprule
Top-5 Acc & \shortstack{HelpSteer2\\val} & \shortstack{LC Alpaca\\Eval 2.0} & \shortstack{Arena-Hard\\v2.0} & \shortstack{Wild\\Bench v2} \\
\midrule
Qwen3-4B     & 0.771 & 0.857 & 0.671 & 0.668 \\
Adapted RLCR & 0.800 & 0.862 & 0.708 & 0.690 \\
\rowcolor{seeblue}
SEE          & \textbf{0.878} & \textbf{0.908} & \textbf{0.748} & \textbf{0.741} \\
\bottomrule
\end{tabular}}
\caption{Top-5 token accuracy: fraction of scoring positions where the judge's score is among the model's five most probable score tokens, averaged over attributes and examples.}
\label{tab:topk}
\end{table}

Table~\ref{tab:topk} shows that the judge's score is sharply localized rather than buried in the tail: SEE places it among the five most probable score tokens 0.8776 of the time on HelpSteer2 validation and 0.9078 on LC AlpacaEval~2.0, exceeding both the base model and Adapted RLCR on every benchmark. This concentration supports downstream use without a judge in the loop.

\begin{figure}[t]
\centering
\includegraphics[width=0.80\columnwidth]{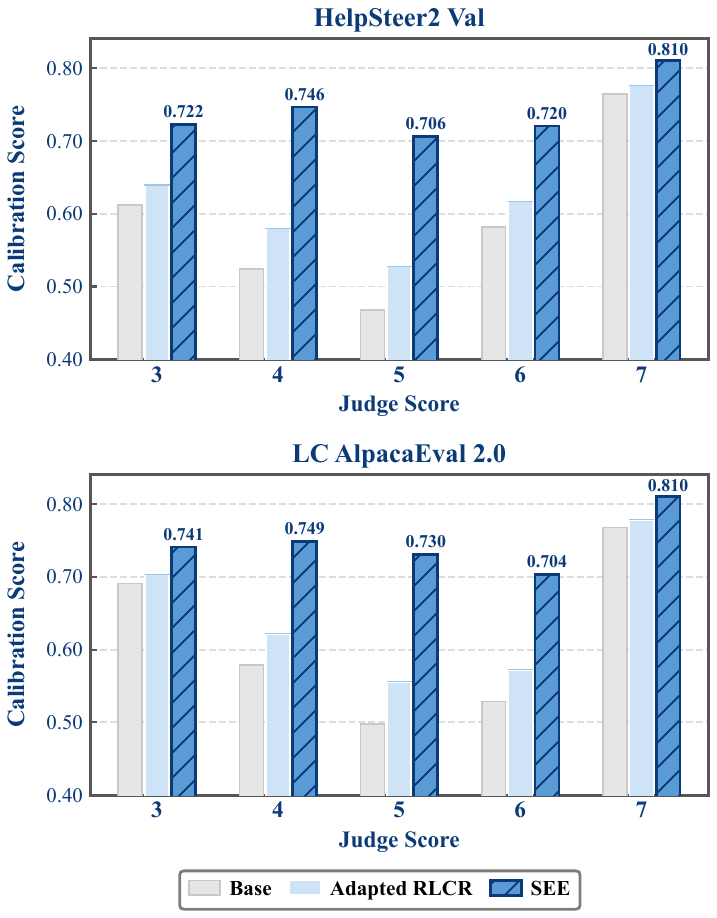}
\caption{Calibration as a function of the judge's score. SEE maintains calibration across the score range, while the base model degrades at the extremes.}
\label{fig:histogram}
\end{figure}

\begin{figure}[t]
\centering
\includegraphics[width=0.95\columnwidth]{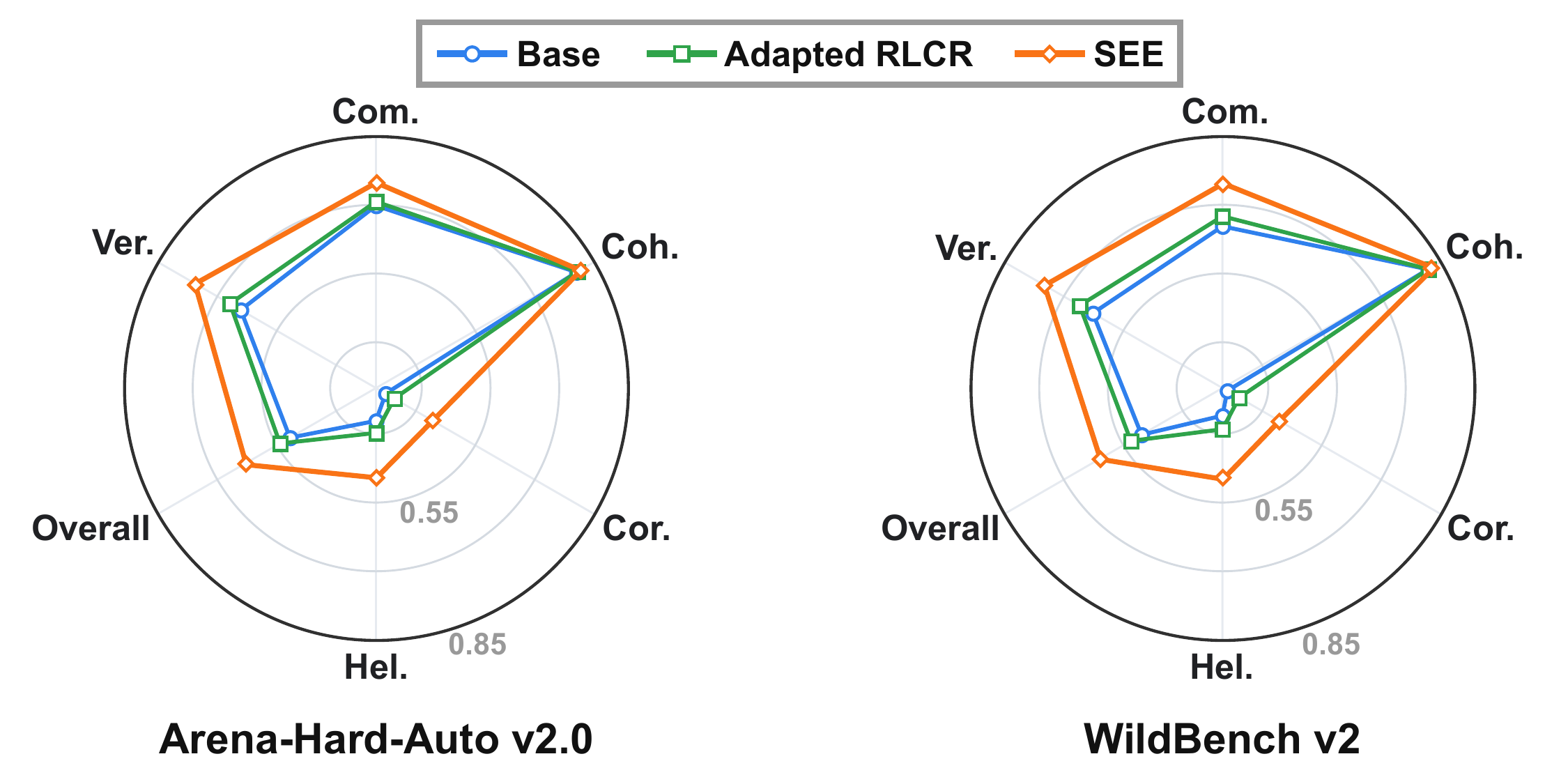}
\caption{Per-attribute calibration on Arena-Hard-Auto~v2.0 and WildBench~v2 for the base model, Adapted RLCR, and SEE. SEE improves every attribute.}
\label{fig:radar}
\end{figure}

\paragraph{Across attributes and score values.}
Figure~\ref{fig:radar} shows that SEE improves over the base model and Adapted RLCR on every attribute, rather than concentrating its gains on a subset, with the largest improvements where the base model starts weakest. Figure~\ref{fig:histogram} provides the complementary score-wise view: the base model handles mid-range judge scores better than the extremes, whereas SEE maintains calibration across the score range. This behavior is consistent with the balanced-coverage resampling used in Masked Judge Distillation.

\section{Discussion}
\label{sec:discussion}

Our central finding is that a base model already approximates an external judge's multi-attribute scores before any targeted training, and that a short cycle is enough to surface this latent ability.
Read together with the broader elicitation literature~\citep{kadavath2022know,zhou2023lima,yue2026does}, this suggests that judge-aligned quality assessment is largely a readout problem: post-training surfaces and enhances it rather than installing it.
That reframing is the main lesson we draw, and it is what turns an expensive training problem into a cheap elicitation one.
The Llama replication preserves the method ordering from a weaker starting calibration, extending SEE to a second model family.

The two phases of SEE improve answer quality and self-evaluation together because their direct supervision falls on different parts of the output.
RL optimizes the entire response, while the distillation phase is confined to the self-evaluation tokens, so the supervised correction sharpens the prediction without perturbing the answer produced by the RL phase.
We see this as a design principle beyond the present setting. When a capability should be added to a model without disturbing an existing one, confining the new supervision to the tokens that carry it, while leaving the rest to a separate objective, keeps the two from competing.

The elicited self-evaluation also appears usable, not merely accurate.
A model that can read its own quality this reliably could, in principle, rerank its own samples or escalate a hard prompt to a stronger model, all without a judge in the loop. This ability is also useful in agentic and embodied settings~\citep{zhang2026don,liu2026palm,yu2026elsa3d,yu2025core3d,li2026toward}.
Establishing how well the elicited self-evaluation drives such decisions is a natural next step, and the localization result makes this plausible.

\section{Conclusion}

We asked whether a base model can predict how an external judge will score its own open-ended responses, and found the ability to be largely present before any targeted training.
SEE elicits it with a short cycle of Calibration-Coupled RL and Masked Judge Distillation, improving held-out calibration across three benchmarks from 160 unique examples while leaving answer quality intact; the elicited self-evaluation is sharply localized in the model's own distribution and stable under judges the model never trained against.
We read this as evidence that judge-aligned self-evaluation is a capability to be surfaced rather than installed, and that the same cyclic recipe may surface other evaluative abilities pretraining has already laid down.

\section*{Limitations}

We evaluate Qwen3-4B and Llama-3-8B with 160 training examples on open-ended, multi-attribute tasks; results may differ at other model scales, data regimes, or task types.
Training labels come from GPT-5.4, with Claude and Gemini as held-out judges. Because all evaluators are LLMs, the results measure cross-judge transfer rather than agreement with human preferences and may inherit judge biases.
Each condition uses one training run, so variability across seeds is not measured.

\section*{Acknowledgments}

This research was funded by the NUS Artificial Intelligence Institute (NAII) seed grant number NAII-SG-2025-025.


\bibliography{references}

\appendix

\section{Training Configuration}
\label{app:config}
 
We report the full training configuration for reproducibility.
Table~\ref{tab:hparams} lists the hyperparameters of the two phases of SEE.
The reinforcement learning phase uses GRPO~\citep{shao2024GRPO}; the distillation phase consists of a single supervised epoch over rollouts selected from the buffer.

\begin{table}[ht]
\centering
\small
\begin{tabular}{ll}
\toprule
\multicolumn{2}{l}{\emph{Calibration-Coupled RL (GRPO)}} \\
\midrule
Rollouts per prompt ($N$)     & 8 \\
Cycles                        & 15 \\
RL steps per cycle            & 10 \\
Batch size                    & 16 \\
Learning rate                 & $1\times10^{-6}$ \\
Optimizer                     & AdamW \\
Weight decay                  & 0.01 \\
KL loss coefficient           & 0.01 \\
Entropy coefficient           & 0.001 \\
Reward weights $(w_q, w_c)$   & $(0.7, 0.3)$ \\
Calibration exponent $\gamma$ & 2 \\
\midrule
\multicolumn{2}{l}{\emph{Masked Judge Distillation (SFT)}} \\
\midrule
\texttt{SFT\_MAX\_SAMPLES}    & 400 \\
Epochs                        & 1 \\
Learning rate                 & $2\times10^{-6}$ \\
Batch size                    & 32 \\
\midrule
\multicolumn{2}{l}{\emph{Shared}} \\
\midrule
Base models                   & \shortstack[l]{\texttt{Qwen3-4B-Base};\\\texttt{Llama-3-8B-Base}} \\
Judge model                   & GPT-5.4 \\
Precision                     & bf16 \\
Max response length           & 8192 \\
Sampling temperature          & 0.8 \\
Sampling top-$p$              & 0.9 \\
Sampling top-$k$              & 25 \\
\bottomrule
\end{tabular}
\caption{SEE training hyperparameters.}
\label{tab:hparams}
\end{table}

\paragraph{SEE versus Adapted RLCR.}
Adapted RLCR is the reinforcement learning phase of SEE run alone, without the distillation phase.
Table~\ref{tab:seevsrlcr} lists the principal settings of the reported runs.
They share the reward, prompts, judge, rollout count, and GRPO optimizer settings; they differ in the distillation phase, batch size, and data budget.
To isolate batch size, we additionally train SEE with batch size 48. The complete matched-batch results are in Appendix~\ref{app:additional:batch}.

\begin{table}[t]
\centering
\small
\begin{tabular}{lcc}
\toprule
& SEE & Adapted RLCR \\
\midrule
Reward (Eq.~\ref{eq:reward})  & same & same \\
Rollouts per prompt ($N$)     & 8 & 8 \\
RL optimizer / LR / KL        & same & same \\
Distillation phase            & yes & no \\
Batch size                    & 16 & 48 \\
Unique examples               & 160 & $\sim$5{,}000 \\
Training epochs / cycles      & 15 cycles & 2 epochs \\
Sample-passes                 & 2{,}400 & $\sim$10{,}000 \\
\bottomrule
\end{tabular}
\caption{Principal settings for SEE and Adapted RLCR. A separate SEE run matches the baseline's batch size of 48 (Appendix~\ref{app:additional:batch}).}
\label{tab:seevsrlcr}
\end{table}

\paragraph{Hardware and infrastructure.}
All runs use four RTX PRO 6000 GPUs (96\,GB each).
We build on VeRL for training and vLLM for rollout generation.

\section{Prompt Templates}
\label{app:prompts}
 
SEE relies on three prompts: the template that instructs the policy to produce
an answer followed by a self-evaluation block, and the system and user prompts
given to the judge.
We reproduce each below.
Placeholders in braces (e.g.\ \texttt{\{REAL\_CONVERSATION\_TRANSCRIPT\}}) are
filled at runtime.
 
\subsection{Self-Evaluation Generation Template}
\label{app:prompts:selfeval}
This template is given to the policy during both phases of training. It elicits
a helpful answer followed by exactly one \texttt{[SELF\_EVAL]} block, and
encodes the format constraints whose violation triggers the penalty in
Equation~\ref{eq:reward}.
 
\begin{lstlisting}[style=prompt]
Write the final assistant output for a conversation.
 
Output requirements:
- Write only the final assistant output. Do not continue the conversation transcript.
- Do not write "User:", "Assistant:", "Conversation:", or "Final assistant output:".
- First answer the latest user request helpfully and safely.
- Then append exactly one self-evaluation block at the very end.
- The self-evaluation block must use exactly these tags and JSON keys:
 
[SELF_EVAL]
{"helpfulness": <integer 0-9>, "correctness": <integer 0-9>, "coherence": <integer 0-9>, "complexity": <integer 0-9>, "verbosity": <integer 0-9>}
[/SELF_EVAL]
 
Scoring guide:
- helpfulness: overall helpfulness of the response to the prompt.
- correctness: inclusion of all pertinent facts without errors.
- coherence: consistency and clarity of expression.
- complexity: intellectual depth required to write the response.
- verbosity: amount of detail relative to what is asked for in the prompt.
 
Important:
- Use integers only: 0-9.
- Always write the self-evaluation block; an otherwise good answer is invalid if it omits the block.
- The answer before [SELF_EVAL] must be non-empty.
- If you use a fenced code block, close it before [SELF_EVAL]; never put the block inside a code fence.
- Do not add text after [/SELF_EVAL]; the final characters of your response must be [/SELF_EVAL].
 
Examples of the required output pattern:
{FEW_SHOT_EXAMPLES}
 
Real conversation:
{REAL_CONVERSATION_TRANSCRIPT}
 
Final assistant output:
\end{lstlisting}
 
\subsection{Judge System Prompt}
\label{app:prompts:judgesys}
The judge scores each response on the five HelpSteer2 attributes. The system
prompt fixes the scale and the rating guidance, including the instruction that
complexity and verbosity are descriptive rather than quantities to maximize.
 
\begin{lstlisting}[style=prompt]
You are an expert evaluator of AI assistant responses. Rate the response on five dimensions using integer scores from 0 to 9.
 
Dimensions:
- helpfulness (0-9): overall helpfulness of the response to the prompt.
- correctness (0-9): inclusion of all pertinent facts without errors.
- coherence (0-9): consistency and clarity of expression.
- complexity (0-9): intellectual depth required to write the response.
- verbosity (0-9): amount of detail in the response, relative to what is asked for.
 
Important rating guidance:
- Higher is not always better for every attribute. Complexity and verbosity are descriptive properties, not targets to maximize.
- For correctness, missing important facts should lower the score even without an obvious false statement.
- For verbosity, judge the amount of detail relative to the user's request.
- For multi-turn conversations, judge the final answer w.r.t. the latest user turn, using the history as context.
- Reference labels use a resized 0-9 scale obtained by doubling the original 0-4 human annotations.
 
When reference responses with human-annotated scores are provided, use them as calibration anchors.
 
Output ONLY a JSON object with five integer scores, nothing else.
\end{lstlisting}
 
\subsection{Judge User Prompt}
\label{app:prompts:judgeuser}
The user prompt presents the conversation and, when available, reference
responses with human-annotated HelpSteer2 scores as calibration anchors. The
target response is passed without its self-evaluation block so the judge scores
the answer alone.
 
\begin{lstlisting}[style=prompt]
[{Role_1}]
{message_1_content}
...
 
Below are reference responses to the same prompt, each with human-annotated scores. Use them as calibration anchors.
 
--- Reference Response 1 ---
{reference_response_1}
Human-Annotated Scores (resized 0-9): {"helpfulness": <int>, "correctness": <int>, "coherence": <int>, "complexity": <int>, "verbosity": <int>}
...
 
[AI Assistant's Response to Evaluate]
{target_response_without_self_eval_block}
 
Rate the response on the resized 0-9 scale. Output ONLY JSON:
{"helpfulness": <int>, "correctness": <int>, "coherence": <int>, "complexity": <int>, "verbosity": <int>}
\end{lstlisting}
\section{Case Studies}
\label{app:cases}

We present two cases from LC AlpacaEval~2.0 that show the full base and SEE
responses together with their scores. Each table reports the model's own
self-evaluation and the judge's scores; SEE rows are shaded. The first case
shows SEE correcting a confident error; the second, more informative case shows
SEE lowering its self-assessment on a merely adequate answer.

\subsection{Correcting a Confident Error}
\label{app:cases:one}

\begin{userbox}
What type of rifle is an AK-47?
\end{userbox}

\begin{baseresp}
The AK47 is a bullpup assault rifle, which means it has the action and magazine
behind the trigger, allowing for a shorter overall length while maintaining a
long barrel. This design was developed by Mikhail Kalashnikov in the Soviet
Union during the 1940s and became one of the most widely used firearms in the
world.
\end{baseresp}

\begin{seeresp}
The AK-47 is a gas-operated, magazine-fed assault rifle. It is widely recognized
for its simplicity, reliability, and robust design. The AK-47 uses a
7.62$\times$39mm cartridge and is known for its durability in various
environments and conditions.
\end{seeresp}

\begin{table}[h]
\centering
\small
\setlength{\tabcolsep}{5pt}
\begin{tabular}{llccccc}
\toprule
Model & Source & Hlp & Cor & Coh & Com & Ver \\
\midrule
Base & Self-eval & 8 & 8 & 8 & 2 & 2 \\
Base & Judge     & 2 & 1 & 8 & 4 & 4 \\
\midrule
\rowcolor{seeblue}
SEE & Self-eval & 8 & 8 & 8 & 3 & 2 \\
\rowcolor{seeblue}
SEE & Judge & 8 & 9 & 9 & 3 & 4 \\
\bottomrule
\end{tabular}
\caption{Case 1 scores. The base model is factually wrong yet rates its own
helpfulness and correctness at 8, while the judge assigns 2 and 1: a confident
error. SEE answers correctly and its self-evaluation closely tracks the judge.}
\label{tab:case1}
\end{table}

The base model describes the AK-47 as a bullpup rifle, which is incorrect, yet
assigns itself high helpfulness and correctness scores; it is not only wrong but
unaware of being wrong. SEE gives a correct, concise answer, and its
self-evaluation aligns closely with the judge on the quality attributes. The
case illustrates SEE reducing confident hallucination and aligning
self-assessment with answer quality.

\subsection{Lowering Confidence on an Adequate Answer}
\label{app:cases:two}

\begin{userbox}
When I open the Anki browser, it shuts down. How can I fix this?
\end{userbox}

\begin{baseresp}
To fix Anki browser shutting down when you open it, try the following steps:
(1)~ensure you have the latest version of Anki installed;
(2)~clear the browser cache and cookies, which can resolve issues with
applications that don't load properly;
(3)~check for conflicts with other browser extensions and disable all extensions
temporarily;
(4)~if the issue persists, uninstall Anki completely, restart, and reinstall the
latest version;
(5)~if none of the above work, reach out to the Anki support team.
\end{baseresp}

\begin{seeresp}
It's possible that the issue is related to the version of Anki you are using or
a conflict with your operating system. Steps to troubleshoot:
(1)~update Anki;
(2)~check compatibility with your operating system;
(3)~run as administrator;
(4)~check for conflicting software;
(5)~clear cache and data;
(6)~check for system updates;
(7)~check for errors in the log;
(8)~reinstall Anki;
(9)~check for hardware issues.
If none of these resolve the issue, contact Anki support or check their forums.
\end{seeresp}

\begin{table}[h]
\centering
\small
\setlength{\tabcolsep}{5pt}
\begin{tabular}{llccccc}
\toprule
Model & Source & Hlp & Cor & Coh & Com & Ver \\
\midrule
Base & Self-eval & 8 & 8 & 8 & 2 & 2 \\
Base & Judge     & 2 & 2 & 8 & 2 & 4 \\
\midrule
\rowcolor{seeblue}
SEE & Self-eval & 5 & 5 & 7 & 4 & 6 \\
\rowcolor{seeblue}
SEE & Judge & 5 & 4 & 8 & 3 & 7 \\
\bottomrule
\end{tabular}
\caption{Case 2 scores. The judge rates both answers as only moderate. The base
model nonetheless self-rates helpfulness and correctness at 8; SEE lowers its
self-assessment to match the judge, the key improvement here.}
\label{tab:case2}
\end{table}

This case is informative precisely because SEE does not earn a high judge score:
the judge rates it as moderate. The base model produces a generic answer with
questionable advice, such as clearing browser cookies and disabling browser
extensions, which is poorly matched to Anki as a desktop application, yet it
self-rates helpfulness and correctness at 8. SEE's answer is also imperfect, but
its self-evaluation is close to the judge's, with helpfulness and correctness
near 5 rather than 8. SEE has learned not to assign high confidence to mediocre
answers, the reliability improvement the calibration results quantify in
aggregate.
\section{Training Algorithm}
\label{app:algorithm}
 
Algorithm~\ref{alg:see} gives the full SEE training procedure. The two phases
alternate for $C$ cycles: Calibration-Coupled RL optimizes the whole response
under the reward of Equation~\ref{eq:reward}, and Masked Judge Distillation
fine-tunes the model on the judge's scores with the loss restricted to the five
self-evaluation score tokens. Only format-valid rollouts enter the buffer, and
the distillation targets are selected by the stratified round-robin of
Section~\ref{sec:method:sft}.
 
\begin{algorithm}[!t]
\caption{Training process of SEE}
\label{alg:see}
\begin{algorithmic}[1]
\Require Base policy $\pi_0$, training set $\mathcal{D}$, cycles $C$, RL steps
per cycle $K$, RL batch size $B_{\text{RL}}$, SFT max samples $M$, SFT batch
size $B_{\text{SFT}}$, judge $J$
\Ensure Final policy $\pi_C$
\State $\pi \gets \pi_0$ \Comment{initialize from the base model}
\For{$c = 1$ to $C$}
  \Statex \textit{// Phase 1: Calibration-Coupled RL}
  \State run GRPO on $\pi$ over $\mathcal{D}$ for $K$ steps, batch size $B_{\text{RL}}$:
  \For{each rollout $y$}
    \State parse answer $a$ and self-eval scores $s$
    \If{format invalid}
      \State $r \gets -1$ \Comment{penalize malformed output}
    \Else
      \State $j \gets J(a)$ \Comment{judge scores the answer}
      \State $q \gets \tfrac{1}{3}\,(j_{\text{hlp}} + j_{\text{cor}} + j_{\text{coh}})/9$
      \State $\mathrm{MAE} \gets \tfrac{1}{5}\sum_{i} |s_i - j_i|$
      \State $\mathrm{cal} \gets (1 - \mathrm{MAE}/9)^{\gamma}$
      \State $r \gets w_q\, q + w_c\, \mathrm{cal}$
    \EndIf
  \EndFor
  \State $\pi_{\text{RL}} \gets$ updated policy
  \State $\mathcal{B} \gets$ format-valid rollouts from this phase \Comment{malformed discarded}
  \Statex \textit{// Phase 2: Masked Judge Distillation}
  \State build targets: replace each rollout's self-eval scores with the judge's scores $j$
  \State select $M$ samples from $\mathcal{B}$ by stratified round-robin over the
  $25$ attribute--score cells (Sec.~\ref{sec:method:sft})
  \State fine-tune $\pi_{\text{RL}}$ on the selected data, batch size
  $B_{\text{SFT}}$, with loss on the five self-eval score tokens only
  \State $\pi \gets \pi_{\text{SFT}}$
\EndFor
\State \Return $\pi$
\end{algorithmic}
\end{algorithm}
\newpage
\section{Detailed Experimental Results}
\label{app:additional}

This appendix provides complete numerical results for the Llama cross-judge evaluation, component ablations, and batch-matched control. Quality (Q), calibration (C), and win-rate (WR) follow Section~\ref{sec:exp:setup}; RLCR abbreviates Adapted RLCR.

\subsection{Llama Cross-Judge Evaluation}
\label{app:additional:crossjudge}

Table~\ref{tab:llama-crossjudge} re-scores the Llama-3-8B models using two judges that supplied no training signal. Quality and calibration preserve SEE $>$ Adapted RLCR $>$ base in all eight judge--benchmark conditions. Win-rate preserves the ordering in seven; the exception is Arena-Hard under Gemini, where SEE trails Adapted RLCR by 0.005.

\subsection{Component Ablations}
\label{app:additional:ablations}

\emph{SEE w/o RL} collects and judges rollouts but performs no RL weight update, so the model changes only through masked judge-score distillation. \emph{SEE w/o strat.} replaces the $5\times5$ stratified round-robin with random buffer selection. \emph{SEE w/o mask} applies distillation loss to all response tokens rather than only the five score tokens. Table~\ref{tab:ablations-full} reports every value plotted in Figure~\ref{fig:ablation-radar}.

\subsection{Batch-Matched Control}
\label{app:additional:batch}

Table~\ref{tab:batch-matched} compares the reported SEE run at RL batch size 16, a retrained SEE run at batch size 48, and Adapted RLCR at batch size 48. The matched run retains SEE's calibration and win-rate advantages, while quality changes by at most 0.006 relative to SEE at batch size 16.

\begin{table*}[p]
\centering
\small
\setlength{\tabcolsep}{3.5pt}
\resizebox{\textwidth}{!}{%
\begin{tabular}{ll cc>{\columncolor{seeblue}}c cc>{\columncolor{seeblue}}c cc>{\columncolor{seeblue}}c}
\toprule
& & \multicolumn{3}{c}{Win-rate} & \multicolumn{3}{c}{Quality} & \multicolumn{3}{c}{Calibration} \\
\cmidrule(lr){3-5}\cmidrule(lr){6-8}\cmidrule(lr){9-11}
Judge & Benchmark & Base & RLCR & SEE & Base & RLCR & SEE & Base & RLCR & SEE \\
\midrule
Claude & HelpSteer2 val       & 0.500 & 0.646 & \textbf{0.748} & 0.412 & 0.492 & \textbf{0.562} & 0.460 & 0.574 & \textbf{0.659} \\
& LC AlpacaEval 2.0    & 0.500 & 0.622 & \textbf{0.673} & 0.510 & 0.594 & \textbf{0.616} & 0.521 & 0.589 & \textbf{0.665} \\
& Arena-Hard-Auto v2.0 & 0.500 & 0.656 & \textbf{0.669} & 0.258 & 0.322 & \textbf{0.352} & 0.345 & 0.384 & \textbf{0.502} \\
& WildBench v2         & 0.500 & 0.681 & \textbf{0.695} & 0.293 & 0.389 & \textbf{0.406} & 0.355 & 0.418 & \textbf{0.534} \\
\midrule
Gemini & HelpSteer2 val       & 0.500 & 0.670 & \textbf{0.749} & 0.529 & 0.652 & \textbf{0.752} & 0.527 & 0.637 & \textbf{0.684} \\
& LC AlpacaEval 2.0    & 0.500 & 0.631 & \textbf{0.664} & 0.685 & 0.764 & \textbf{0.773} & 0.575 & 0.619 & \textbf{0.697} \\
& Arena-Hard-Auto v2.0 & 0.500 & \textbf{0.557} & 0.552 & 0.365 & 0.402 & \textbf{0.411} & 0.362 & 0.411 & \textbf{0.559} \\
& WildBench v2         & 0.500 & 0.608 & \textbf{0.626} & 0.393 & 0.491 & \textbf{0.509} & 0.390 & 0.457 & \textbf{0.604} \\
\bottomrule
\end{tabular}}
\caption{Llama-3-8B cross-judge results. Both judges are held out from training. Bold marks the best value within each metric and row.}
\label{tab:llama-crossjudge}

\vspace{1.25em}
\setlength{\tabcolsep}{2.5pt}
\resizebox{\textwidth}{!}{%
\begin{tabular}{l >{\columncolor{seeblue}}c >{\columncolor{seeblue}}c >{\columncolor{seeblue}}c ccc ccc ccc}
\toprule
& \multicolumn{3}{c}{SEE} & \multicolumn{3}{c}{w/o RL} & \multicolumn{3}{c}{w/o strat.} & \multicolumn{3}{c}{w/o mask} \\
\cmidrule(lr){2-4}\cmidrule(lr){5-7}\cmidrule(lr){8-10}\cmidrule(lr){11-13}
Benchmark & Q & C & WR & Q & C & WR & Q & C & WR & Q & C & WR \\
\midrule
HelpSteer2 val       & 0.704 & 0.731 & 0.671 & 0.642 & 0.696 & 0.504 & 0.688 & 0.707 & 0.624 & 0.652 & 0.674 & 0.527 \\
LC AlpacaEval 2.0    & 0.792 & 0.746 & 0.592 & 0.711 & 0.713 & 0.495 & 0.793 & 0.722 & 0.579 & 0.767 & 0.705 & 0.502 \\
Arena-Hard-Auto v2.0 & 0.575 & 0.609 & 0.518 & 0.472 & 0.606 & 0.470 & 0.571 & 0.600 & 0.511 & 0.507 & 0.583 & 0.482 \\
WildBench v2         & 0.567 & 0.609 & 0.581 & 0.464 & 0.601 & 0.473 & 0.562 & 0.589 & 0.546 & 0.499 & 0.579 & 0.488 \\
\bottomrule
\end{tabular}}
\caption{Full Qwen3-4B component ablations. HelpSteer2 WR is quality score win-rate; the open-ended benchmarks use response win-rate.}
\label{tab:ablations-full}

\vspace{1.25em}
\setlength{\tabcolsep}{6pt}
\begin{tabular}{l ccc >{\columncolor{seeblue}}c >{\columncolor{seeblue}}c >{\columncolor{seeblue}}c ccc}
\toprule
& \multicolumn{3}{c}{SEE-16} & \multicolumn{3}{c}{SEE-48 (batch matched)} & \multicolumn{3}{c}{Adapted RLCR-48} \\
\cmidrule(lr){2-4}\cmidrule(lr){5-7}\cmidrule(lr){8-10}
Benchmark & Q & C & WR & Q & C & WR & Q & C & WR \\
\midrule
HelpSteer2 val & 0.704 & 0.731 & 0.671 & 0.701 & 0.722 & 0.673 & 0.662 & 0.675 & 0.570 \\
LC AlpacaEval 2.0 & 0.792 & 0.746 & 0.592 & 0.786 & 0.741 & 0.588 & 0.789 & 0.716 & 0.534 \\
Arena-Hard-Auto v2.0 & 0.575 & 0.609 & 0.518 & 0.577 & 0.591 & 0.522 & 0.569 & 0.547 & 0.502 \\
WildBench v2 & 0.567 & 0.609 & 0.581 & 0.561 & 0.587 & 0.569 & 0.561 & 0.541 & 0.531 \\
\bottomrule
\end{tabular}
\caption{Batch-matched Qwen3-4B control. The matched SEE-48 columns are shaded. Q, C, and WR denote quality, calibration, and win-rate.}
\label{tab:batch-matched}
\end{table*}

\clearpage

\section{Responsible Research Details}
\label{app:responsible}

\paragraph{Artifacts, licenses, and intended use.}
The main artifacts used in the study are \texttt{Qwen3-4B-Base}, \texttt{Llama-3-8B-Base}, HelpSteer2, LC AlpacaEval~2.0, Arena-Hard-Auto~v2.0, WildBench~v2, GRPO, VeRL, vLLM, and proprietary LLM judges.
We cite the creators of the model, datasets, benchmarks, methods, and software in Sections~\ref{sec:exp:setup} and~\ref{sec:method}, and in Appendix~\ref{app:config}.
The \texttt{Qwen3-4B-Base} model card lists Apache-2.0 terms; the Llama checkpoint is used under the Llama 3 community license; HelpSteer2 is released under CC-BY-4.0; and the AlpacaEval, Arena-Hard-Auto, and WildBench code repositories list Apache-2.0 licenses.
The proprietary judges are accessed only through their provider APIs and are subject to the corresponding provider terms.
We use these artifacts for research on model evaluation and post-training, which is consistent with their role as open model, dataset, benchmark, and evaluation artifacts.
We do not redistribute model weights, raw benchmark data, proprietary judge models, or API response logs in the paper package.
The accompanying software archive is intended to document and reproduce the training procedure; users must obtain external models, datasets, and API access under their own applicable licenses and terms.

\paragraph{Data content and privacy.}
We do not collect new user data or recruit new annotators.
The training data are derived from HelpSteer2, and the evaluation prompts come from existing public instruction-following benchmarks.
Because some benchmarks contain open-ended real or crowdsourced user prompts, they may contain sensitive, offensive, or identifying content inherited from the source artifacts.
We do not attempt to identify users, infer protected attributes, or release raw prompts and responses beyond the short qualitative examples in Appendix~\ref{app:cases}; those displayed examples were manually inspected for obvious identifying information.
The released code package excludes raw datasets, generated outputs, rollout logs, API-key files, and experiment logs.

\paragraph{Potential risks of misuse.}
SEE is a general post-training method for teaching a model to anticipate an evaluator's scores.
Although we study benign response-quality assessment, the same mechanism could be paired with an evaluator that rewards harmful behavior or used to help a model anticipate automated oversight, potentially supporting evaluator gaming or more capable harmful outputs.
Our release contains the training procedure but excludes harmful datasets, generated outputs, proprietary judge access, and experiment logs; deployments in sensitive domains should retain independent oversight and appropriate safeguards.

\paragraph{Data and split statistics.}
SEE trains on the first 160 unique prompts in the HelpSteer2 training split's dataset order: we scan from the beginning, discard repeated prompts, and retain the first 160 that remain, with no filtering by attribute, score, topic, or length.
These prompts are reused across 15 cycles for 2{,}400 total sample-passes for both base-model families.
The Adapted RLCR baseline trains for two epochs over roughly 5{,}000 unique examples, or about 10{,}000 sample-passes.
We evaluate on HelpSteer2 validation and on LC AlpacaEval~2.0, Arena-Hard-Auto~v2.0, and WildBench~v2, as described in Section~\ref{sec:exp:setup}.

\paragraph{Compute and implementation details.}
The base models have 4B and 8B parameters.
All reported training runs use four RTX PRO 6000 GPUs with 96\,GB memory each, bf16 precision, VeRL for GRPO training, and vLLM for rollout generation.
Appendix~\ref{app:config} lists the optimizer, batch sizes, learning rates, reward weights, sampling parameters, and other hyperparameters.
The full experimental study, including the Llama replication, component ablations, and matched-batch control, used more than 300 GPU-hours in total.
The software archive includes the package versions used by the released training scripts.

\paragraph{Result aggregation.}
Reported quality, calibration, win-rate, and top-5 token accuracy values are aggregate means over the relevant evaluation examples from one training run per condition.

\paragraph{AI assistance.}
AI assistants were used for writing, editing, and code/documentation support. All scientific claims, experiments, analyses, and final text were verified and edited by the authors.

\paragraph{Code availability.}
We provide a GitHub repository for the training pipeline at
\href{https://github.com/YiShan05/SEE_official}{\nolinkurl{github.com/YiShan05/SEE_official}}.
The repository contains the core SEE implementation, including data preparation,
Calibration-Coupled RL, rollout collection, score-token SFT construction, and
Masked Judge Distillation scripts.

\end{document}